%% file: main.tex
\documentclass[journal]{IEEEtran}
\usepackage{cite}
\usepackage{amsmath,amssymb,amsfonts}
\usepackage{algorithmic}
\usepackage{graphicx}
\usepackage{algorithm,algorithmic}
\usepackage{hyperref}
\usepackage[permil]{overpic} 
\usepackage{booktabs}       
\usepackage{dsfont}
\usepackage{nicefrac}       
\usepackage{microtype}      
\usepackage{adjustbox}
\usepackage{makecell}
\usepackage{multirow}
\usepackage{array}
\usepackage{yhmath}
\usepackage{stmaryrd}
\usepackage{yfonts}
\usepackage{tabularx}
\usepackage{textcomp}
\usepackage{subcaption}
\usepackage{float}
\usepackage{ragged2e}
\usepackage{multirow}
\usepackage{color, colortbl}
\usepackage{comment}
\usepackage[T1]{fontenc}

\def\y{\checkmark}
\DeclareMathOperator*{\argmax}{arg\,max}
\definecolor{Lblue}{rgb}{0.69,0.92,0.95}
\definecolor{gray}{gray}{0.85}
\hypersetup{hidelinks=true}
\usepackage{textcomp}
\def\BibTeX{{\rm B\kern-.05em{\sc i\kern-.025em b}\kern-.08em
    T\kern-.1667em\lower.7ex\hbox{E}\kern-.125emX}}

\begin{document}
\title{Wound3DAssist: A Practical Framework for \\ 3D Wound Assessment}
\author{
Remi Chierchia, 
Rodrigo Santa Cruz,
L\'eo Lebrat,
Yulia Arzhaeva,
Mohammad Ali Armin, 
Jeremy Oorloff,
\\Chuong Nguyen,
Olivier Salvado,
Clinton Fookes, 
and
David Ahmedt-Aristizabal
\thanks{This work was supported by the MRFF Rapid Applied Research Translation grant (RARUR000158), CSIRO AI4M Minimising Antimicrobial Resistance Mission, and Australian Government Training Research Program (AGRTP) Scholarship.}
\thanks{Remi Chierchia, David Ahmedt-Aristizabal, Rodrigo Santa Cruz, Yulia Arzhaeva, Mohammad Ali Armin, Jeremy Oorloff, and Chuong Nguyen are with the Imaging and Computer Vision Group, Data61, CSIRO (e-mail: remi.chierchia@data61.csiro.au; david.ahmedtaristizabal@data61.csiro.au; rodrigo.santacruz@csiro.au; yulia.arzhaeva@data61.csiro.au; ali.armin@data61.csiro.au; jeremy.oorloff@data61.csiro.au; chuong.nguyen@data61.csiro.au).}
\thanks{Remi Chierchia, L\'eo Lebrat, Olivier Salvado, Clinton Fookes are with the School of Electrical Engineering \& Robotics, Queensland University of Technology, Australia (e-mail: remi.chierchia@hdr.qut.edu.au; leo.lebrat@qut.edu.au; olivier.salvado@qut.edu.au; c.fookes@qut.edu.au).}
}

\maketitle

\begin{abstract}
Managing chronic wounds remains a major healthcare challenge, with clinical assessment often relying on subjective and time-consuming manual documentation methods. Although 2D digital videometry frameworks have aided wound measurement, these approaches struggle with perspective distortion, a limited field of view, and an inability to capture wound depth, especially in anatomically complex or curved regions. To overcome these limitations, we present Wound3DAssist, a practical framework for 3D wound assessment using monocular consumer-grade videos.
Our framework generates 3D wound models from short handheld recordings captured using consumer-grade devices, 
enabling non-contact, automatic measurements from reconstructed multi-view surfaces. We integrate 3D reconstruction, wound segmentation, tissue classification, and periwound analysis into a modular workflow. We evaluate Wound3DAssist across digital models with known geometry, silicone phantoms, and real patients. 
Results show that the framework supports high-quality wound bed visualization, approximately millimeter-scale surface reconstruction accuracy in the evaluated clinical cases, and multi-view wound-tissue composition analysis. Full assessments are completed in under 20 minutes, demonstrating feasibility for a research framework intended for future clinical workflow evaluation.
\end{abstract}


\section{Introduction}

\IEEEPARstart{C}{hronic}
wounds pose a significant clinical and economic burden, affecting millions of patients worldwide and costing healthcare systems billions annually~\cite{nussbaum2018economic,sen2021human}. Accurate and consistent wound assessment is essential for diagnosis, monitoring, and treatment planning~\cite{chang2011comparison}. Traditional wound documentation methods include contact-based manual measurements and 2D digital videometry, and remain affected by operator variability, perspective ambiguity, and lack of depth perception~\cite{filko2018wound}. These challenges hinder longitudinal tracking and objective comparison of wound healing over time.
Recent advances in computer vision and 3D reconstruction have opened the door to more precise and standardized wound analysis~\cite{souto2023three,niri2021multi}. Three-dimensional (3D) data provide richer geometric context, enabling accurate wound area, perimeter, and depth measurements---critical metrics for evaluating wound progression~\cite{pena2020evaluation}. Moreover, 3D models allow better visualization and communication among clinicians, improving documentation and decision-making.

Several research efforts have explored the application of 3D imaging to wound care, employing technologies such as structured light~\cite{zahia2020integrating,zhang2023rgb}, multi-spectral~\cite{juszczyk2020wound,chang2017multimodal}, and photogrammetry~\cite{malian2005development,wannous2010enhanced}.
While promising, these works often remain limited in scope or usability. Many rely on proprietary hardware not publicly available or require extensive post-processing and technical expertise, making them impractical for routine clinical application. Critically, there is also a lack of standardized datasets for benchmarking, and few existing works provide an end-to-end pipeline that integrates 3D reconstruction, tissue segmentation, and automated wound measurement~\cite{wannous2010enhanced,chang2017multimodal,filko2025autonomous,souto2023three,sirazitdinova2017system} into a single solution.

\begin{figure*}[!t]
    \centering
    \begin{overpic}[width=0.95\linewidth]{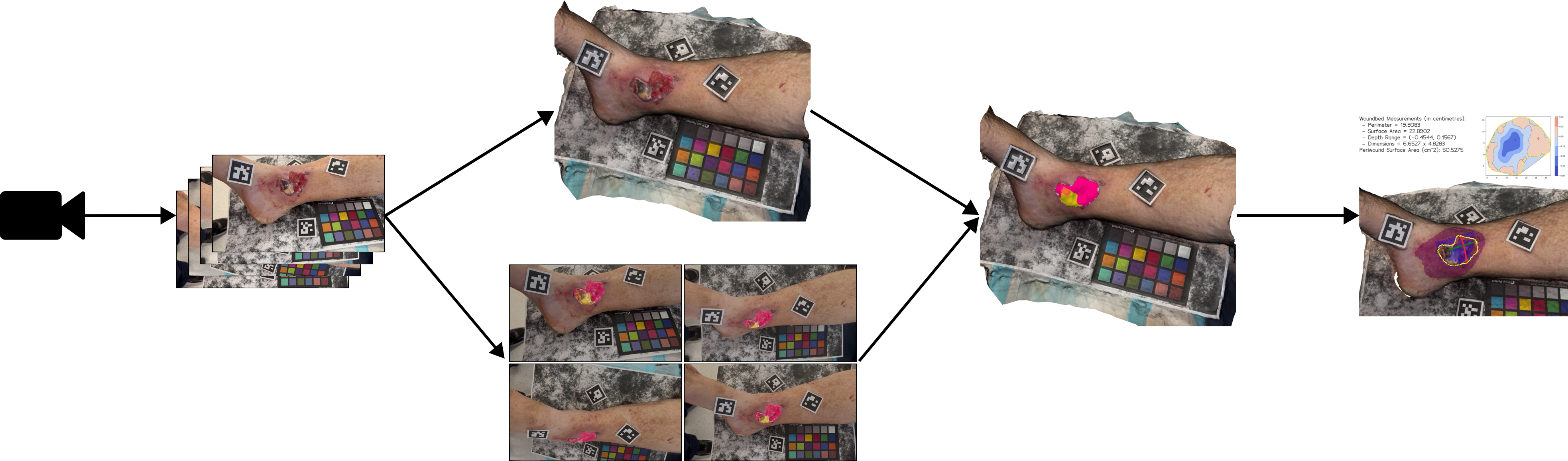}
        \put (95,85) {\small \parbox{3cm}{\centering Frame Extraction\\ \& Selection}}
        \put (370,136) {\small 3D Reconstruction}
        \put (345,300) {\small \textit{Wound Bed Visualization}}
        \put (305,-16) {\small 2D Segmentation \& Post-processing}
        \put (625,59) {\small \parbox{3cm}{\centering2D-to-3D Segmentation Mapping}}
        \put (615,233) {\small \textit{Wound Tissue assessment}}
        \put (851,68) {\small \parbox{3cm}{\centering3D Wound\\ Documentation}}
        \put (850,240) {\small \parbox{3cm}{\centering\textit{Wound Bed \& Periwound assessment}}}
    \end{overpic}
    \vspace{10pt}
    \caption{Main components and capabilities of our proposed wound documentation framework \textbf{Wound3DAssist}:
    \textit{Frame Extraction \& Selection} (Sec.~\ref{sec:meth_frame_sel}); 
    \textit{3D Reconstruction} (Sec.~\ref{sec:3d_recon}) supporting interactive wound bed visualization; 
    \textit{2D Segmentation \& Post-processing} (Sec.~\ref{sec:2d_segmentaiton}); 
    \textit{2D-to-3D Segmentation Mapping} (Sec.~\ref{sec:2d-3dseg}) enabling automated wound tissue assessment; and 
    \textit{3D Wound documentation} (Sec.~\ref{sec:woundbed_model}) reporting comprehensive wound bed and periwound assessment.}
    \label{fig:syste}
\end{figure*}

To address these gaps, we introduce Wound3DAssist, an end-to-end 3D wound assessment framework utilizing consumer-grade video (Table~\ref{tab:system_comp}). 
We advance prior work through five contributions: 
(1) an end-to-end workflow integrating 3D reconstruction, segmentation, and wound documentation, evaluated using real clinical recordings; 
(2) a quality-aware frame-selection procedure that reduces the influence of motion-blurred frames before reconstruction; 
(3) a geometry-aware 2D-to-3D aggregation method that maps semantic labels using view-dependent weighting; 
(4) a surface-cover-based measurement method for deriving curvature-aware wound dimensions and depth estimates; and 
(5) a three-tier evaluation spanning digital wounds, silicone phantoms, and prospective clinical data.
Furthermore, as displayed in Fig.~\ref{fig:syste}, the framework supports four core wound assessments: i) \textit{interactive 3D wound bed visualization}; ii) \textit{automated wound bed segmentation and geometric assessment}; iii) \textit{wound bed tissue classification}; and iv) \textit{geometric periwound segmentation and surface-area assessment}.

%
By combining consumer-grade acquisition devices with a modular computational pipeline, Wound3DAssist provides a research framework for investigating 3D wound documentation under realistic acquisition conditions. The evaluation and discussion also identify current limitations related to dataset availability, imaging artefacts, cross-modality reference data, and clinical validation.


\section{Experimental Design}
A critical step in validating 3D wound assessment frameworks is access to diverse, well-characterized datasets that reflect both controlled and real-world conditions. However, 3D wound datasets with ground truth annotations are scarce and often lack the realism or variability needed to test model robustness across clinical contexts. Collecting 3D data from actual wounds presents additional difficulties: it is time-consuming, may cause patient discomfort, and can disrupt clinical workflows---particularly when aiming to capture a broad range of wound types, skin tones, lighting conditions, and patient positions.
Existing publicly available wound datasets are commonly tailored for 2D analysis of single-frame images~\cite{thomas2020medetec,wang2020fully,wang2021foot,kendrick2022translating,pereira2022complexwounddb}, with few allowing for 3D imaging~\cite{juszczyk2020wound,krkecichwost2021chronic}, synthetic modeling~\cite{sinha2024dermsynth3d}, or medical phantoms such as Symour II~\footnote{https://www.mentone-educational.com.au/health-education/health-education/seymour-ii-wound-care-model}.

To overcome these limitations, we design a multi-tiered evaluation strategy using: 
(i) a digital wounds dataset with known 3D geometry for quantitative evaluation, 
(ii) a high-resolution phantom wound dataset captured under controlled conditions, 
and (iii) a new clinical dataset featuring real-world wound recordings.
Each dataset uniquely evaluates the framework’s components, gradually increasing scene complexity at development stages.
%
We also define a set of metrics aligned with clinical wound documentation practices to evaluate the framework's performance across the datasets.

\vspace{-6pt}
\subsection{Digital wounds dataset \label{sec:data_digital}}


Synthetic datasets offer precise control over scene complexity, simulating various lighting conditions, skin tones, and camera trajectories. Direct access to ground truth geometry and segmentation enables systematic testing of reconstruction accuracy
and robustness under idealised conditions. In this direction, we use a synthetic dataset combining 3D human body avatars, real wound textures, and parametrically defined wound geometries. For detailed methodology and results, we refer the reader to~\cite{lebratISBI24}.

\vspace{-6pt}
\subsection{Silicone wounds dataset}\label{sec:silicone_dataset}


Silicone phantoms are realistic replicas of clinical wounds, developed by companies such as TraumaSIM~\footnote{https://traumasim.com.au/} to assist the training of healthcare professionals. 
We simulate a typical clinical scenario where an operator acquires handheld recordings in a controlled environment, introducing practical challenges such as multi-light reflections and motion blur.
This allows us to evaluate the framework's robustness to variations in camera quality and capture conditions, not present in the synthetic dataset.
Furthermore, repeated acquisitions can be evaluated for precision without the ethical constraints typically encountered in clinical settings. For detailed methodology and dataset specifications, we refer the reader to~\cite{chierchia2024salve}.

\input{tables/system_comparison}

\vspace{-6pt}
\subsection{Clinical wounds dataset}
\label{sec:real_wound_dataseet}

We evaluate the framework in realistic clinical settings by collecting two complementary datasets.
The retrospective dataset consists of pre-existing annotated 2D wound images, whereas the prospective dataset comprises multi-view recordings captured by trained clinical staff using a structured acquisition protocol.
These datasets contain real challenges encountered in clinical assessments, including complex wound topologies, heterogeneous patient conditions, and involuntary patient motion.


\subsubsection{Retrospective data}
\begin{figure}[!t]
    \centering
    \includegraphics[width=0.98\linewidth]{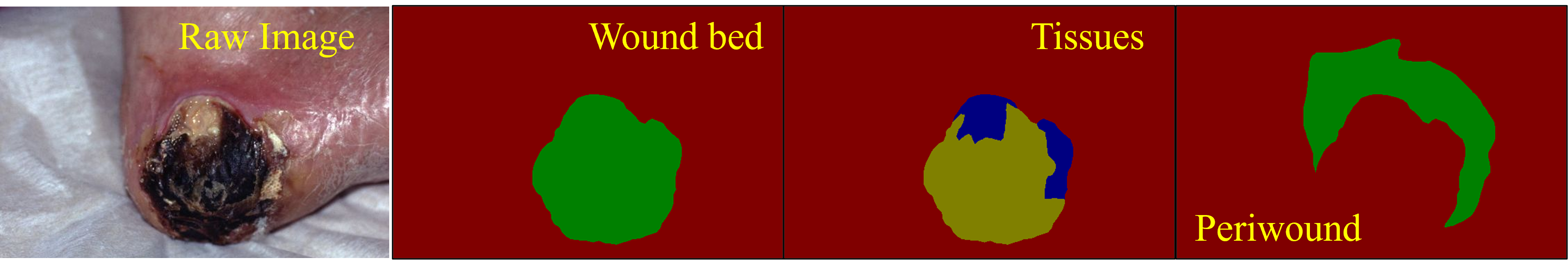}
    \caption{Sample retrospective image and its corresponding wound bed, tissue (necrotic and slough), and periwound annotations.}
    \label{fig:retrospective}
\vspace{-0.5cm}
\end{figure}

The retrospective dataset comprises 2D wound images captured in simulated clinical settings and made available under a data-use agreement. It was used exclusively to train and internally evaluate the 2D segmentation models supporting the subsequent 2D-to-3D pipeline. The task-specific image sets were randomly divided using an 80/20 training and internal-evaluation split; no retrospective images were used for 3D reconstruction or wound measurement.
The dataset includes wound-bed, periwound, and tissue annotations created in CVAT by trained wound-care experts, as illustrated in Fig.~\ref{fig:retrospective}. 
Tissue annotations are restricted to the wound bed, whereas the periwound region represents the surrounding skin outside the wound-bed boundary. The dataset predominantly contains ulcerative wounds but also includes pressure injuries, dermatitis, skin tears, skin cancer, surgical and traumatic wounds, burns, and skin inflammation. In total, 829 images contain wound-bed annotations, including 592 with periwound annotations and 710 with tissue annotations. Across all tissue-annotated pixels, granulation tissue accounts for 43.48\%, slough for 32.45\%, necrotic tissue for 21.98\%, and epithelial tissue for 2.09\%.

\label{sec:prospective_data}

\subsubsection{Prospective data} 
The prospective dataset was collected in collaboration with clinical nurse researchers using a structured acquisition protocol designed to support multi-view 3D reconstruction. Operators were instructed to follow a gantry-style trajectory consisting of a smooth sweep over and around the visible wound to provide overlapping views and reduce abrupt camera motion. Videos were recorded using iPhones and Logitech cameras. Revopoint scans were acquired as reference geometry where available, and ArUco markers were positioned near the wound to enable evaluation and scale recovery. A formal usability study was not conducted, and the required operator-training time was not quantified.
Unlike the retrospective dataset, the prospective dataset was reserved entirely as an unseen evaluation set for Wound3DAssist. It comprises 57 annotated images organised into 19 wound-recording entries from 10 patients. Because some entries contain multiple wounds and others correspond to repeat visits, the 19 entries do not represent 19 distinct wounds. Across these entries, 21 wound occurrences were recorded: 16 ulcers—nine venous, four diabetic, and three arterial—two dermatitis or hyperkeratosis occurrences, one surgical wound, one traumatic bite or puncture wound, and one wound classified as other. Two trajectories were selected for longitudinal analysis.
The same CVAT-based annotation protocol as for the retrospective dataset was applied. Wound experts annotated the wound bed, periwound region, and granulation, slough, necrotic, and epithelial tissue. These tissues were present in 23, 37, 3, and 2 images, respectively. Each prospective image was annotated by a single expert; therefore, inter-rater reliability could not be assessed. These annotations served as the evaluation reference.

\section{Existing Frameworks} 

3D analysis frameworks enable computation of depth and volume, key metrics for comprehensive wound assessment~\cite{pena2020evaluation,joel2021Reliability}. However, challenges remain due to the quality of acquisition devices, environmental variables, and algorithm robustness~\cite{joel2021Reliability,chierchia2024salve}. 
In the literature, there are two major approaches to recovering depth information: multi-modal camera approaches, which combine RGB with specialised sensors; and photogrammetry-based approaches, which employ single or multiple color (RGB) cameras.
In Table~\ref{tab:system_comp}, we report our framework's contributions compared to existing works.

\noindent\textbf{Multi-modal camera approaches.}
Pavlov{\v{c}}i{\v{c}} et al.~\cite{pavlovvcivc2015wound} introduced a swing arm system using both a laser projector and an RGB camera to capture 3D wound surface profiles.
Juszczyk et al.~\cite{juszczyk2020wound} used a tripod-mounted rig combining depth, stereo, thermal, and RGB cameras, with Poisson surface reconstruction from both depth and stereo cameras.
Chang et al.~\cite{chang2017multimodal} proposed a more sophisticated probe integrating thermal/infrared, multi-spectral, and chemical sensors; the wound surface was obtained via convex hulls on a fused RGB-depth point cloud. All the above methods require manual wound delineation.
Filko et al.~\cite{filko2025autonomous} employed a robotic arm equipped with RGB, depth, and a 3D scanner guided by SegFormer-based~\cite{xie2021segformer} networks for wound detection and tissue classification. 
Although precise, these systems are expensive and require specialized training to operate them. 
Filko et al.~\cite{filko2018wound} evaluated three different RGB-D (capturing color and depth images) consumer-grade cameras, which proved accurate on the wound model Saymour II.
Zahia et al.~\cite{zahia2020integrating} integrated a Structure Sensor (commercial product) with an iPad and MaskRCNN for wound segmentation. 
Zhang et al.~\cite{zhang2023rgb} proposed a lightweight, portable prototype with an Intel RealSense RGB-D camera, Raspberry Pi, and touch screen. For automatic segmentation of the wound, the authors developed a model based on an encoder-decoder architecture. 
These methods primarily rely on single-frame input, which may limit surface coverage for large wounds or wounds located on curved anatomical regions.

\noindent\textbf{Photogrammetry-based approaches.}
Early works implemented a projector to illuminate the wound with a pattern acquired with one~\cite{plassmann1998mavis} or three~\cite{malian2005development} cameras. Those systems were bulky, making them impractical for use in hospitals, clinics, and aged care facilities.
Wannous et al.~\cite{wannous2010enhanced} proposed a low-cost two-view setup using photogrammetry
and traditional classification techniques (SVM).
A two-view segmentation was merged onto the 3D by recursively splitting the mesh triangles into sub-triangles in the presence of different classes. Unlike one view, it improved robustness against medical reference.
%
Several studies have leveraged Structure-from-Motion (SfM) from videos or multi-view images. Mirzaalian et al.~\cite{mirzaalian2019measuring} used video and a Random Forest for segmentation within a predefined ROI, and obtained more accurate wound area measurements compared to the 2D approach.
%
Liu et al.~\cite{liu2019wound} used a smartphone to capture multiple images for 3D reconstruction and 2D-based interactive segmentation, demonstrating improved accuracy in curved regions.
%
S{\'a}nchez-Jim{\'e}nez et al.~\cite{sanchez2022sfm} proposed 3DULC, software to generate orthographic views from SfM outputs for manual tracing and automatic area estimation.
%
More recent systems have integrated learning-based segmentation. Souto et al.~\cite{souto2023three} used U-Nets 
for wound segmentation and classification. The wound region was then masked to extract the 3D model via OpenMVS, a recent SfM-based software. 
%
Niri et al.~\cite{niri2021multi} employed Meshroom, the most recent SfM pipeline among the previous methods, using dual U-Nets to segment wound and skin regions, followed by multiple 3D segmentation algorithms. Re-projecting the segmentation to 2D improved upon the initial estimations, highlighting data augmentation capabilities.
Inspired by the flexibility of photogrammetry approaches that do not require complex acquisition systems, we adopt monocular video as our input.
We devise a modular framework that supports interchangeable 3D reconstruction and segmentation pipelines. 
Our prior SALVE study~\cite{chierchia2024salve} compared photogrammetric and neural reconstruction methods in terms of surface fidelity, inter-device and inter-recording consistency, computational requirements, and processing time.

\section{Wound3DAssist Framework}
As illustrated in Fig.~\ref{fig:syste}, the workflow begins with capturing a video of the wound using a consumer-grade device. From this video, a set of key frames is automatically selected based on wound visibility, marker presence, and image sharpness (Section~\ref{sec:meth_frame_sel}).
The selected frames are processed by the photogrammetric reconstruction pipeline (Section~\ref{sec:3d_recon}), which estimates camera poses and generates a textured 3D mesh.
The identical frame subset is then processed by the 2D segmentation models, and the resulting predictions are mapped onto the mesh using the estimated camera poses. This produces a segmented 3D wound model (Section~\ref{sec:3d_segm}) that enables automated wound tissue assessment.
Subsequently, the framework computes wound metrics directly from the segmented mesh (Section~\ref{sec:woundbed_model}). Fiducial markers are used to retrieve physical scale (Section~\ref{sec:met_scale}) to convert all measurements into metric units. 
Together, these components enable comprehensive wound bed and periwound assessment.

\vspace{-6pt}
\subsection{3D wound reconstruction}

\subsubsection{Pre-processing}
\label{sec:meth_frame_sel}

To create a high-quality 3D wound model, the framework selects approximately $N = 50$ sharp frames that capture the wound from multiple viewpoints.
Following the data curation protocol established in the SALVE dataset~\cite{chierchia2024salve}, we utilize the variance of the Laplacian as a quantitative sharpness indicator to automatically filter out motion-blurred images. The same selected subset is used for reconstruction, 2D segmentation, and 2D-to-3D mapping.
%
Reliable 3D wound reconstruction requires a relatively static scene, an adequately textured surface, and overlapping camera angles. While our frame selection process successfully filters out motion blur, the system cannot overcome severe glare, occlusions, sudden lighting changes, or a lack of sufficient overlap.

\subsubsection{3D reconstruction methods}
\label{sec:3d_recon}
Currently, our framework adopts the Meshroom pipeline~\cite{alicevision2021}, a photogrammetry-based method for 3D reconstruction.
Specifically, we applied two custom settings optimized for our dataset:
i) \textit{Single-Camera Model}: Tailored for monocular video input;
ii) \textit{High-Density SIFT Features}: Up to 50,000 SIFT features per image to improve detail, with a minor increase in processing time.
Using these matches, the pipeline estimates camera parameters and produces a sparse 3D point cloud via Structure-from-Motion (SfM). 
Subsequently, a Multi-View Stereo (MVS)~\cite{hartley2003multiple} step generates a dense point cloud from which the textured mesh is produced.

%
%
%
To validate our selection of the Meshroom pipeline over other approaches, we rely on our prior benchmarking study (SALVE~\cite{chierchia2024salve}). In that work, we quantitatively compared standard photogrammetry against modern neural rendering methods (e.g., NeRFs and 3D Gaussian Splatting). While neural methods achieved marginal gains in surface fidelity, they incurred greater computational costs unsuitable for rapid clinical workflows. Therefore, our well-configured Meshroom pipeline~\cite{alicevision2021} was selected as it offers the optimal practical balance, delivering sub-millimeter accuracy while maintaining robustness and hardware efficiency.


\vspace{-6pt}
\subsection{2D-to-3D segmentation} 
\label{sec:3d_segm}

%
While prior studies in both medical domains~\cite{niri2021multi} and general stereo imaging~\cite{li2020learning} demonstrate that multi-view aggregation improves upon single-view methods, evaluating direct 3D segmentation models is highly challenging without 3D ground-truth data.
To address this, we adopt a modular 2D-to-3D segmentation approach. This strategy not only facilitates seamless integration with diverse 3D reconstruction methods, but it also enables evaluation by reprojecting the aggregated 3D labels into held-out expert-annotated 2D views.

\begin{figure}[!t]
     \centering
     \begin{overpic}[width=\linewidth]{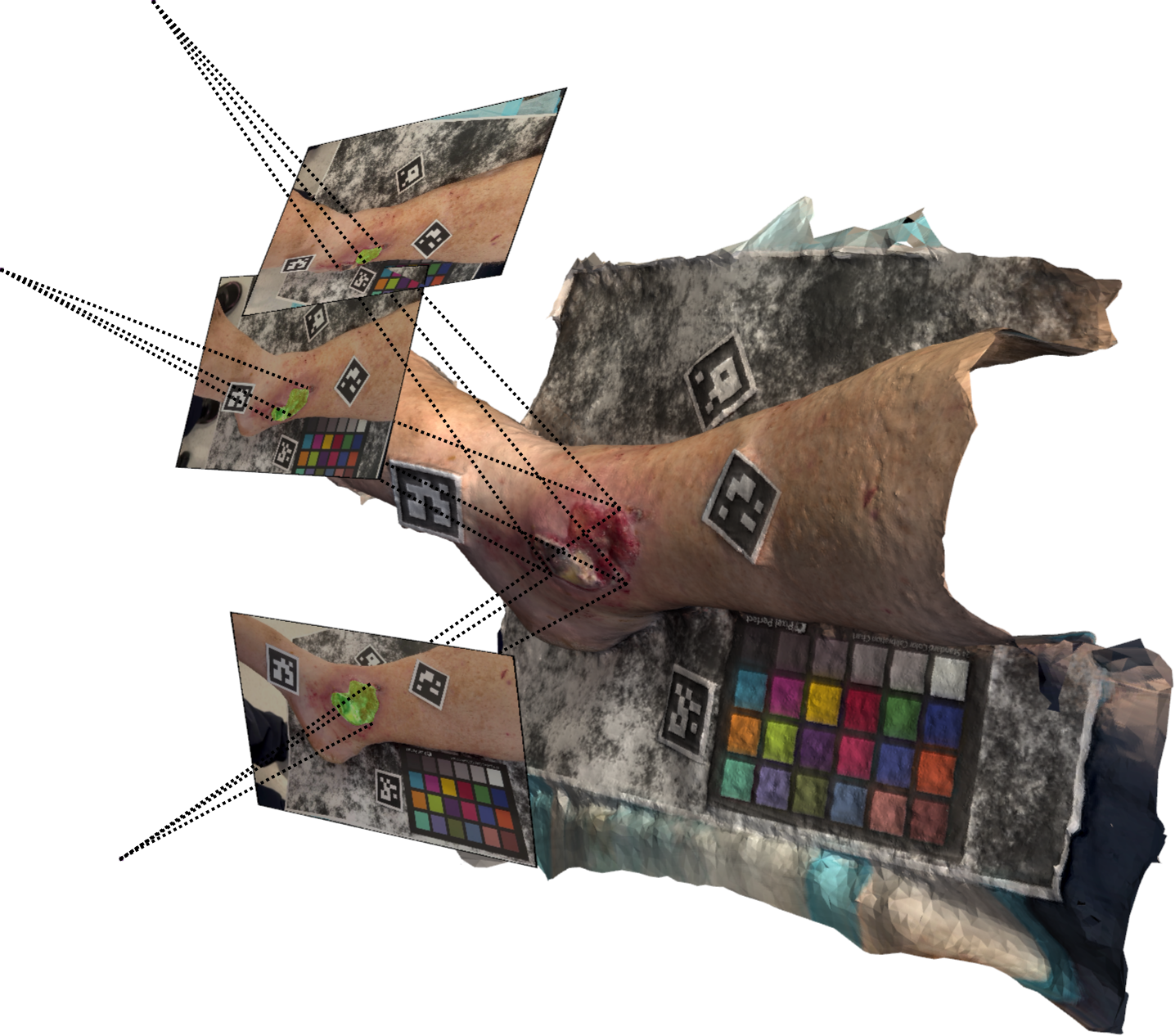}
        \put (150,150) {\small $\pi^i_{f_j}$}
    \end{overpic}
     \caption{Visual representation of the 2D-to-3D Segmentation Mapping. Each ray within the same camera $i$ will map to a different reconstructed mesh face $f_j$.}
     \label{fig:3D_segm_diagram_new}
\vspace{-6pt}
\end{figure}

\subsubsection{2D segmentation}
\label{sec:2d_segmentaiton}

While various architectures have been proposed for precise medical image segmentation, including hybrid dual-path U-Nets~\cite{ahmed2025d2u}, our method builds upon the SegFormer architecture~\cite{xie2021segformer}, specifically the MiT-b5 variant, which we fine-tuned to predict 2D segmentations of the wound bed, periwound region, and four tissue classes (granulation, necrotic, slough, and epithelial).
Specifically, we fine-tuned the model for three distinct tasks: binary segmentation of the wound bed, binary segmentation of the periwound region, and multi-class segmentation for the five tissue categories (inclusive of the background class).

The task-specific retrospective image sets were randomly divided using an 80/20 training and internal-evaluation split. To address pixel-level class imbalance, we used cross-entropy loss weighted according to the inverse frequency of each label, assigning greater weight to under-represented tissue classes. No class-specific oversampling or undersampling was applied. Training augmentations included resizing to $512\times512$, random horizontal flipping, random blurring, random distortions, and normalisation. The prospective dataset was not used for training or model selection.


\subsubsection{2D-to-3D segmentation mapping}
\label{sec:2d-3dseg}

To obtain a 3D segmentation, we employ a rasterization algorithm~\cite{pineda1988rasterization} to project the 3D reconstruction mesh onto the predicted 2D segmentation masks. For each available 2D view, we determine vertex visibility and assign a label by aggregating predictions using a majority voting scheme. 
The diagram of this approach is depicted in Fig.~\ref{fig:3D_segm_diagram_new}.
%
%
%
Additionally, the voting includes a weighting factor to reflect the reliability of each candidate label. 
%
We assume that 2D segmentations are most reliable when the wound surface is approximately perpendicular to the camera's viewing direction. 
Given $N$ cameras, we first compute a weighting factor defined by:
\vspace{-6pt}
\begin{equation}
    w_j^i = \left| \langle \mathbf{n}_{f_j}, \pi^i_{f_j}\rangle \right|,
\end{equation}
where $\mathbf{n}_{f_j}$ is the normal of the reconstructed mesh face $f_j$, and $\pi^i_{f_j}$ is the normalized ray from the camera $i$ intersecting the face barycenter.
Then, for each face, we compute its segmentation label $L$ through majority voting~\cite{Bai2024Stability}. More specifically, we compute the solution of the following maximization problem:
\vspace{-6pt}
\begin{equation}\label{eq:wmv}
L_j = \argmax \sum_{i=1}^N \mathds{1}_{w_j^i \geq  1/2} \log\left( \frac{w_j^i}{1 - w_j^i} \right)\mathbf{p}_i \,,
\end{equation}
where $\mathbf{p}_i \in \{0,1\}^{C}$ is the argmax of the probability vector from the prediction of image $i$ associated with the camera ray $\pi^i_{f_j}$.
To ensure reliability, we only include projections where $w_j^i \geq  \cos(\pi/3)$, thereby excluding views that are too oblique with the wound surface.
Note that for periwound segmentation, we compute Eq.~\ref{eq:wmv} for all values of $w_j^i$ as this region is often undersampled due to the recordings focusing primarily on the wound bed.
Finally, the assigned labels can be optionally post-processed with binary morphology operations applied to a mesh as a graph~\cite{Najman2014graphmorph}, to remove small labeled ``islands'' and refine the contours of the segmented regions. 

\vspace{-6pt}
\subsection{3D wound documentation}
\label{sec:woundbed_model}
The final component of our framework performs 3D wound bed analysis on the reconstructed mesh and its corresponding segmentation.

We reorient the reconstructed mesh to a \textit{reference coordinate frame} by translating the wound bed's centroid to the origin and rotating its principal axes computed through Principal Component Analysis (PCA) to match a Euclidean coordinate system. The average normal vector of the wound cavity is used to enforce $z$-upwards orientation, providing a consistent coordinate convention for the subsequent measurements.

We compute a \textit{wound bed surface cover}, a virtual reference surface fitted to the wound-bed perimeter and intended to approximate the surrounding intact skin surface. It provides a consistent computational reference for estimating wound length, width, and depth.
%
We fit a smooth surface $f(x, y)$ to the wound bed perimeter vertices using Radial Basis Function (RBF) interpolation with a Thin-Plate Spline (TPS) kernel. 
\begin{align}\label{eq:cover}
f(x,y) &= \sum_{i=1}^{n} w_i \phi(\| (x, y) - (x_i, y_i) \|),\\
&\text{where } \phi(r) = r^2 \log r. \nonumber
\end{align}
\\
Given a set of $n$ 3D-points ${(x_i, y_i, z_i)}_{i=(1,\dots, n)}$ located on the perimeter of the wound, the surface is obtained by solving
\begin{equation}
\min_{w_i} \sum_{i=1}^{n} \| w_i \phi(\| (x, y) - (x_i, y_i) \|) - z_i \|.
\end{equation}
In the following sections, we describe how to derive 3D measurements with further details.

\subsubsection{Wound bed perimeter}

To compute the perimeter of the wound bed, we first isolate individual wounds by extracting the connected components from the segmented mesh. Wound boundaries are extracted by identifying mesh edges incident to a single triangle; the vertices of these edges define the wound perimeter.
We fit a periodic $\alpha$-smooth B-spline curve $\gamma_\alpha(s)$ parameterized over the interval $s \in [0,1]$, to the wound boundary vertices. The resulting arc length provides an estimate of the wound perimeter:
\begin{equation}
P = \int_0^1 \left| \frac{d\gamma_{\alpha}(s)}{ds} \right| ds,
\end{equation}
%
The choice of $\alpha$ (smoothness regularization) should be adapted to the wound type to ensure accurate and robust perimeter estimation.

\subsubsection{Wound bed length and width}

The length and width represent the two largest dimensions of the wound bed. 
Let $P = \{ v_1, v_2, \dots, v_{|P|} \}$ be the set of wound bed perimeter vertices. 
The wound length is defined as the maximum distance between any pair of perimeter vertices:
\begin{equation}
(v^l_{\text{start}}, v^l_{\text{end}}) = \underset{v_i,v_j \in P}{\arg\max}~D(v_i, v_j),
\end{equation}
where $D(v_i, v_j)$ denotes the approximated geodesic distance computed over the \textit{wound bed surface cover} between $v_i$ and $v_j$. 
To compute this, we discretize the Euclidean line segment connecting $v_i$ and $v_j$ into a set of intermediate points ${ p_k }$. These points are projected onto the fitted surface cover using the interpolated function $f$ of Eq.~\ref{eq:cover}.
A B-spline curve $\gamma(s)$ is then fitted through the projected points, and the approximated geodesic distance is obtained by computing its arc length:
\begin{equation}\label{eq:distance}
D(v_i, v_j) = \int_0^1 \left| \frac{d\gamma(s)}{ds} \right| ds.
\end{equation}

To compute the wound width, we search for perimeter vertex pairs that form an approximately orthogonal direction to the wound length segment $(v^l_{\text{start}}, v^l_{\text{end}})$. Among all pairs $(v_i, v_j) \in P \times P$ satisfying the angular constraint $\cos\left(\theta(v^l_\text{start}v^l_\text{end},v^l_iv^l_j)\right) \leq \varepsilon$, we select the pair with the longest geodesic distance computed as in Eq.~\ref{eq:distance}.

\begin{figure}[!t]
    \centering
    \begin{overpic}[width=\linewidth]{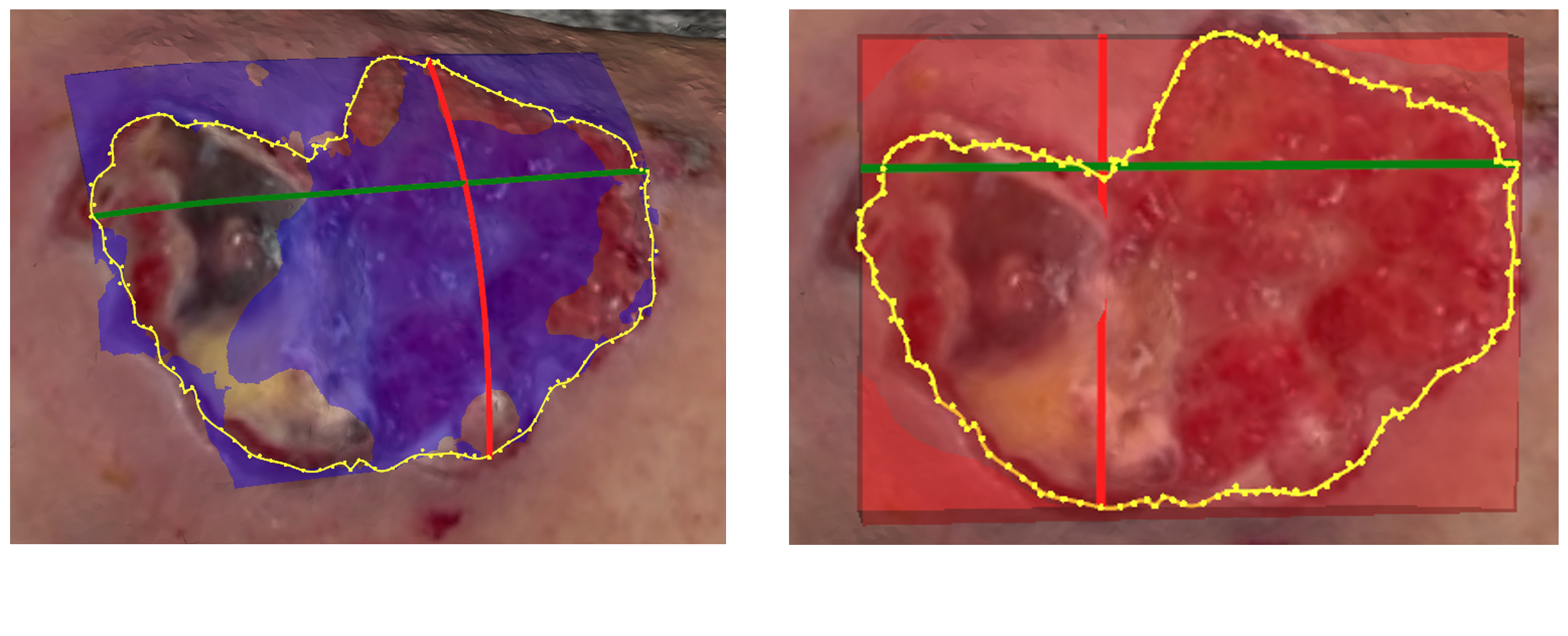} 
        \put (100,20) {\small Surface cover path}
        \put (650,20) {\small Bounding box}
    \end{overpic}
    \caption{Measuring wound length and width along the fitted surface cover (left) accounts for local curvature, unlike a global bounding-box measurement (right).}
    \label{fig:mea_lw}
\vspace{-6pt}
\end{figure}

As illustrated in Fig.~\ref{fig:mea_lw}, this approach mimics the clinical practice of using a paper ruler to assess wound dimensions. 
Unlike a global minimum enclosing box, this method follows the fitted surface curvature and therefore provides curvature-aware wound dimensions.

\subsubsection{Wound bed and periwound surface area}

The wound bed and periwound areas are computed by summing the areas of triangular mesh faces assigned to each respective segmentation label in the reconstructed surface.
%
This produces surface-based area estimates directly from the labelled triangular mesh faces. 
The periwound analysis is limited to expert-defined spatial segmentation and geometric surface-area estimation; it does not assess erythema, oedema, temperature, texture, infection, or other diagnostic characteristics.

\subsubsection{Wound bed depth}

\begin{figure}[!t]
    \centering
    \includegraphics[width=0.95\linewidth]{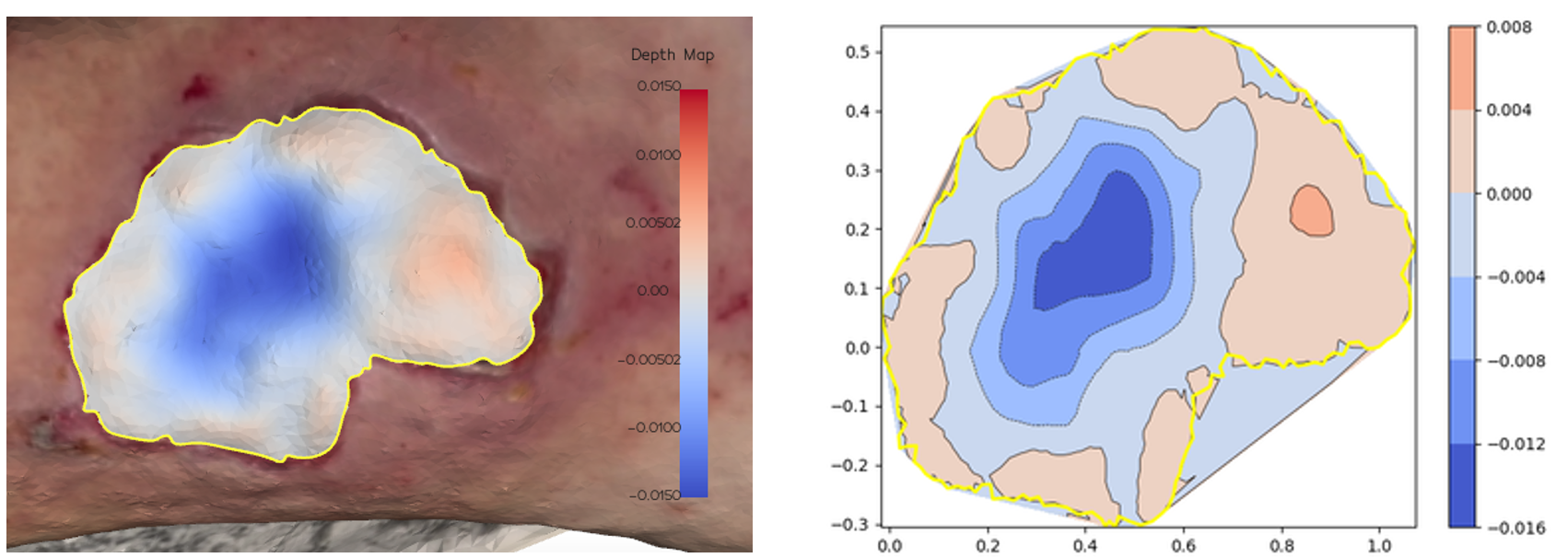}
    \caption{Estimated wound-depth values relative to the fitted surface cover, visualised on the 3D mesh (left) and as a topographic map (right).}
    \label{fig:meas_depth_map}
\vspace{-6pt}
\end{figure}

Given the reoriented wound bed and its corresponding surface cover, the depth at a vertex is defined as the vertical distance along the $z$-axis between the vertex and its projection onto the surface cover.
A positive depth indicates a \textit{protrusion} of the wound bed above the surface cover, while a negative depth corresponds to a \textit{depression} below the surface.

As illustrated in Fig.~\ref{fig:meas_depth_map}, these depth values can be visualized directly on the 3D reconstructed mesh by applying a color map to the wound bed vertices. Alternatively, they can be rendered as a topographic 2D map or summarized using statistical measures such as the maximum wound depth.
These values are depth estimates relative to the computed surface cover. Direct patient-level ground truth for maximum wound depth was not available in the present study.

\subsubsection{Wound bed tissue composition}

To analyze the tissue composition of the wound bed, we compute the surface area covered by each tissue type.
The percentage of coverage is then the ratio to the total wound bed area.
%
This provides a quantitative description of the estimated tissue distribution and supports comparison across recorded visits.

\subsubsection{Scale retrieval} 
\label{sec:met_scale}

A well-known limitation of 3D reconstruction from images is that the resulting model is defined only up to a similarity transformation relative to the real-world object~\cite{hartley2003multiple}. 
We use ArUco markers placed near the wound and captured in the video sequence. The markers provide a known physical reference that enables scale recovery. 
By detecting and extracting the 2D-pixel coordinates of its corners, we can triangulate its 3D coordinates using the Direct Linear Transformation (DLT)~\cite{hartley2003multiple} algorithm.
Solving across multiple frames yields robust 3D estimates of the marker corners.
We then compute the distances between adjacent corners to estimate the average side length of the marker in the reconstructed model, deriving the scaling factor by comparison.

\section{Results and Evaluation}

\subsection{3D reconstruction evaluation}\label{sec:eval_3d_proc}

To evaluate 3D reconstruction quality, we first align the reconstruction and reference meshes using the Iterative Closest Point (ICP) algorithm. Then, we sample both meshes uniformly to enable point-wise comparison.
Similarly to~\cite{chierchia2024salve}, we define a polygon around the wound in both the reconstructed mesh and the ground-truth point cloud. Across the three datasets, we computed Chamfer Distance (CD), Absolute Distance (AD), Hausdorff Distance (HD), and Normal Consistency (NC).

\subsubsection{Results on digital wounds dataset}
\begin{figure}[!t]
    \centering
    \begin{overpic}[width=\linewidth]{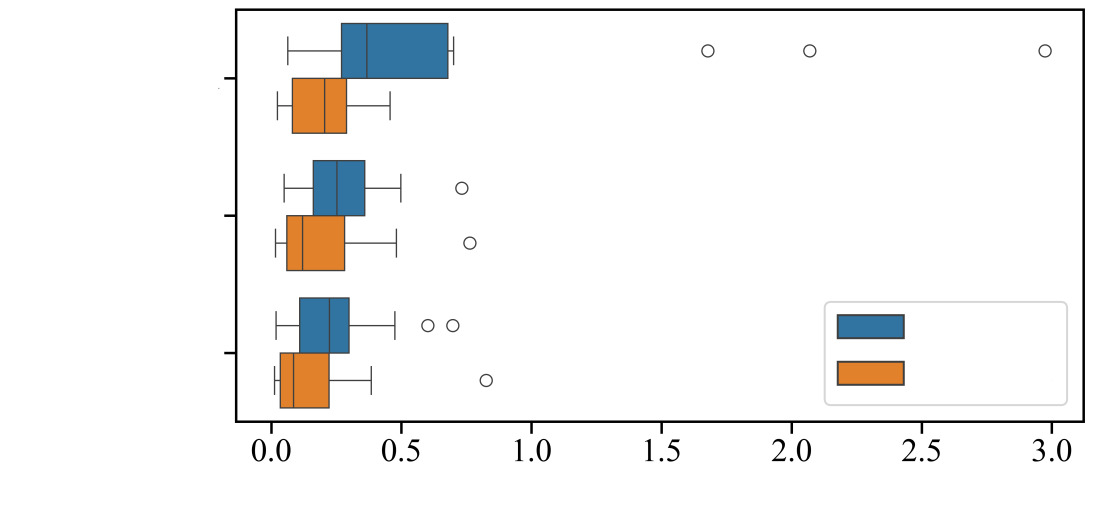}
        \put (510,20) {\footnotesize Chamfer (mm)}
        \put (22,393) {\footnotesize Gaussian blur}
        \put (45,267) {\footnotesize Motion blur}
        \put (-3,141) {\footnotesize No perturbation}
        \put (840,163) {\footnotesize Subject 1}
        \put (840,120) {\footnotesize Subject 2}
    \end{overpic}
    \caption{Assessing the impact of camera motion on mesh quality. The impact of Gaussian blur and motion blur artifacts is consistent across the digital wounds dataset of each subject.}
    \label{fig:3d_eval_digital_blur}
\end{figure}

We evaluated reconstruction accuracy using the digital wound dataset (Section~\ref{sec:data_digital}), which provides ground-truth geometry for objective assessment.
Higher-resolution images consistently improved reconstruction quality, while increasing frame count offered diminishing returns.
We also analyzed the impact of common artifacts (e.g., shadows, reflections) and simulated variations in camera motion to assess robustness (Fig.~\ref{fig:3d_eval_digital_blur}). 
These experiments informed the acquisition and frame-selection protocol but do not establish uniform robustness across all clinical imaging conditions.

\subsubsection{Results on silicone wounds dataset}
On the silicone dataset (Section~\ref{sec:silicone_dataset}), we evaluated the geometric accuracy and precision of several reconstruction methods across recording devices. 
%
These results supported the selection of Meshroom as a practical balance of reconstruction quality and computational cost for the current implementation.
The complete quantitative benchmarking is reported in~\cite{chierchia2024salve}.

\subsubsection{Results on clinical wounds dataset}

We performed a quantitative analysis using wound recordings collected from actual patients in healthcare settings (as described in Section~\ref{sec:real_wound_dataseet}). 
For evaluation purposes, we focused on a subset of cases for which both iPhone video recordings and 3D scans from a Revopoint scanner were available. 
These included: \texttt{P1-01}, \texttt{P1-02}, \texttt{P2-01}, \texttt{P2-02}.
All recordings were captured under consistent conditions, with no changes to the wound scene or marker placement between acquisitions. 

\begin{figure}[!t]
    \centering
    \begin{overpic}[width=\linewidth]{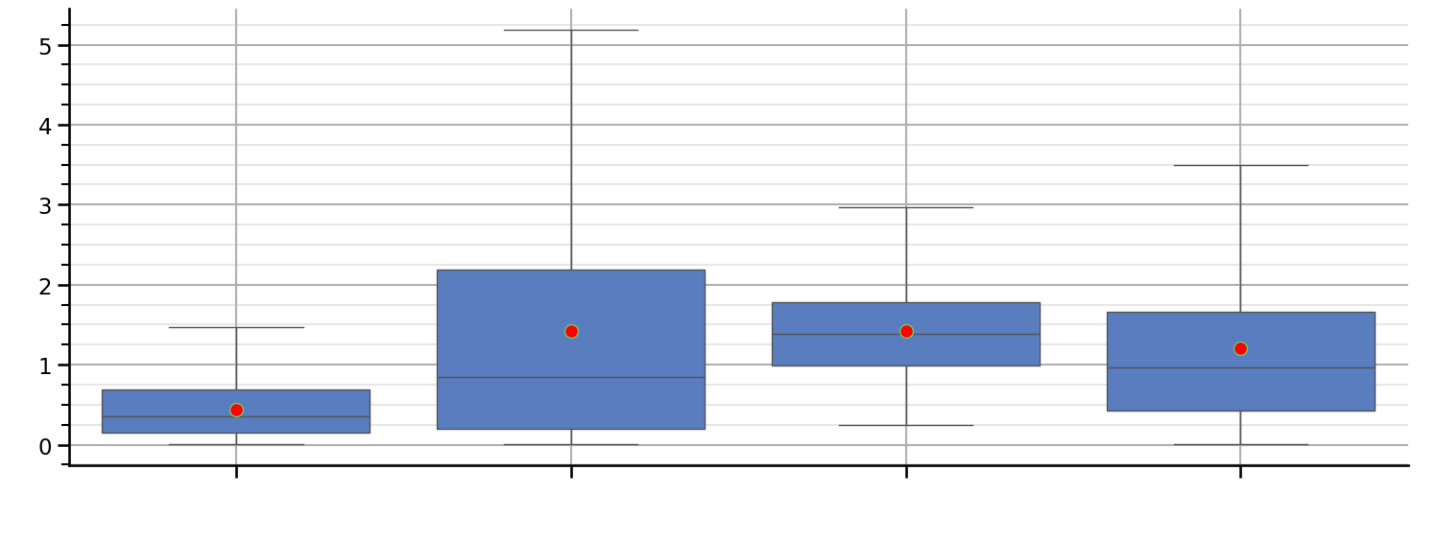} 
        \put (-10,140) {\rotatebox{90}{\footnotesize \text{AD (mm)}}}
        \put (117,15) {\footnotesize \texttt{P2-02}}
        \put (350,15) {\footnotesize \texttt{P1-01}}
        \put (583,15) {\footnotesize \texttt{P2-01}}
        \put (815,15) {\footnotesize \texttt{P1-02}}
    \end{overpic}
    \caption{Point-wise distances between the Revopoint reference point clouds and the manually aligned reconstructions for the four evaluated clinical cases.}
    \label{fig:3d_recon_real_boxplot}
\vspace{-6pt}
\end{figure}

\begin{figure}[!t]
    \centering
    \begin{overpic}[width=\linewidth]{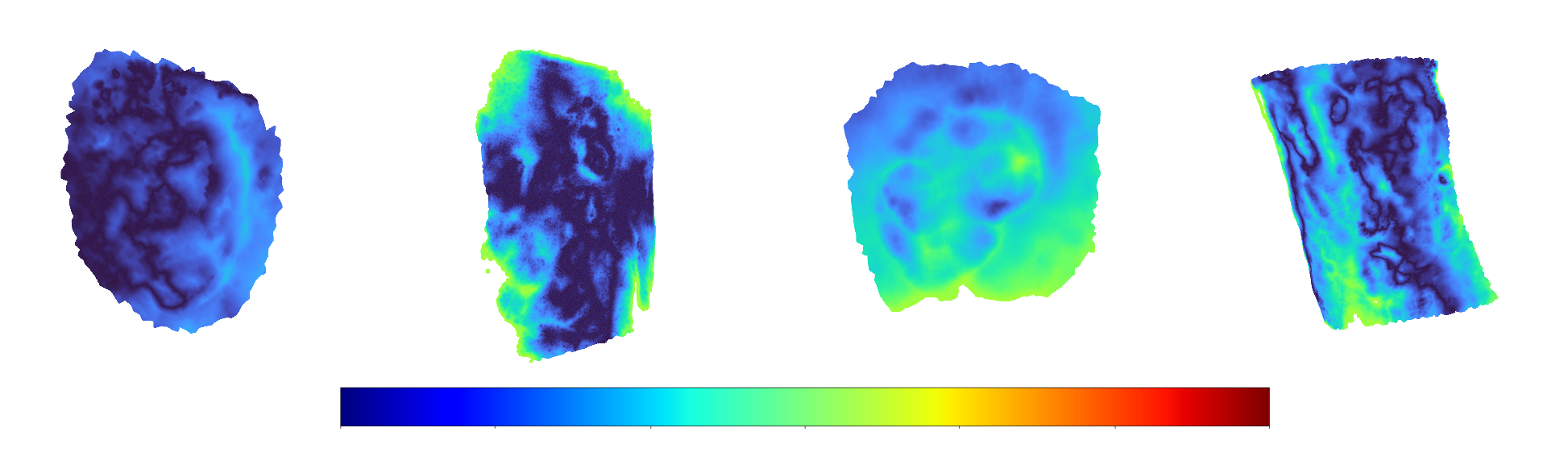}
        \put (65,270) {\footnotesize \texttt{P2-02}}
        \put (315,270) {\footnotesize \texttt{P1-01}}
        \put (580,270) {\footnotesize \texttt{P2-01}}
        \put (820,270) {\footnotesize \texttt{P1-02}}

        \put (205, 5) {\tiny 0.00}
        \put (304, 5) {\tiny 0.83}
        \put (403, 5) {\tiny 1.67}
        \put (501, 5) {\tiny 2.50}
        \put (600, 5) {\tiny 3.33}
        \put (699, 5) {\tiny 4.17}
        \put (797, 5) {\tiny 5.00}
        \put (475, 55) {\tiny AD (mm)}
    \end{overpic}
    \caption{Color-coded maps showing the spatial distribution of reconstruction errors for real wound data. The errors represent the point-wise distances across the wound surfaces.}
    \label{fig:3d_recon_real_map}
\vspace{-6pt}
\end{figure}

\begin{figure}[!t]
    \centering
    \begin{overpic}[width=\linewidth]{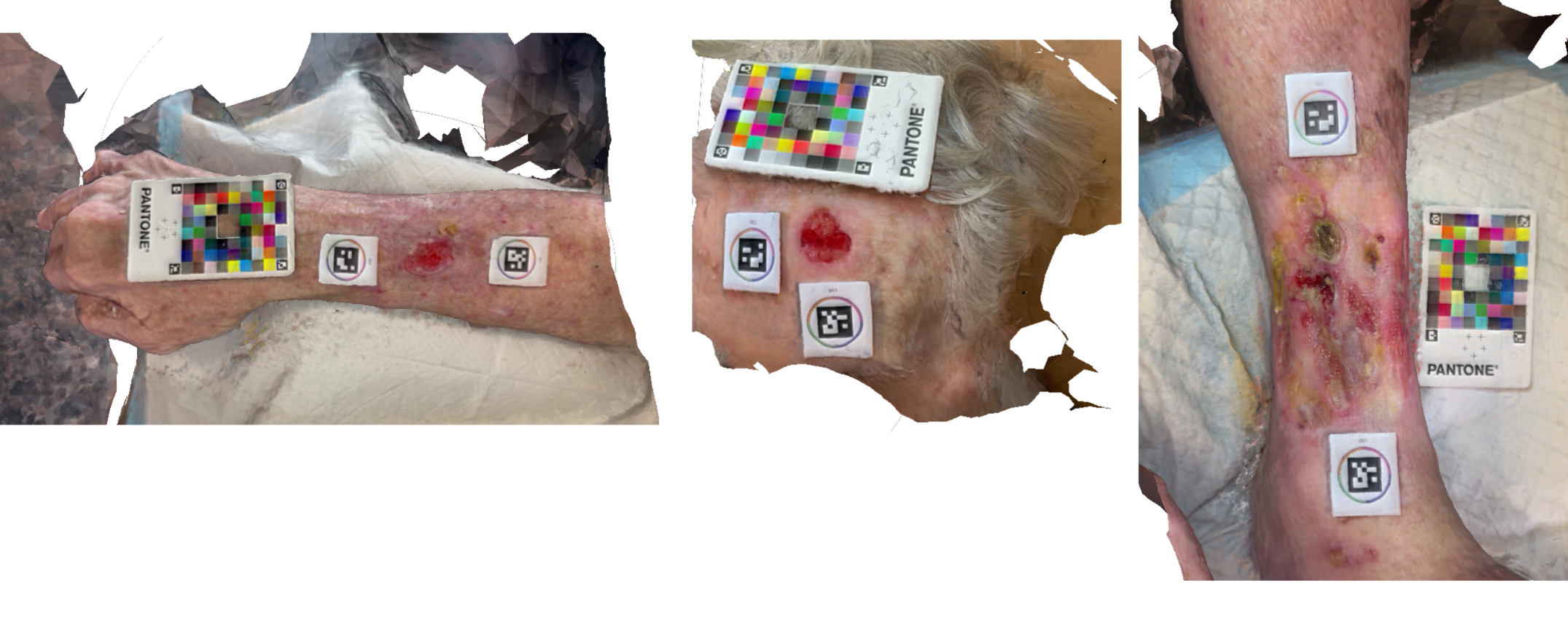}
        \put (180,100) {\footnotesize \texttt{P2-02}}
        \put (530,100) {\footnotesize \texttt{P2-01}}
        \put (820,0) {\footnotesize \texttt{P1-02}}
    \end{overpic}
    \caption{Examples of 3D reconstructions from clinical data illustrating our method's ability to capture detailed wound geometry and surface structure under realistic conditions.}
    \label{fig:3d_recon_real_samples}
\vspace{-6pt}
\end{figure}

Because automatic alignment was not feasible, the reconstructed meshes were manually aligned with the Revopoint reference point clouds. This procedure may introduce alignment-related error and should therefore be considered when interpreting the results. Across the four evaluated cases, the reconstructions achieved an average point-wise surface error of approximately 1 mm (Fig.~\ref{fig:3d_recon_real_boxplot}). This value characterises overall surface-reconstruction fidelity and does not constitute a direct validation of maximum wound depth or volume.
Most discrepancies were concentrated near wound bed boundaries, where irregular geometry and limited texture make reconstruction more difficult. Some higher errors in peripheral regions were also associated with evaluation-time cropping differences (Fig.~\ref{fig:3d_recon_real_map}).
These findings provide preliminary evidence that the framework can recover clinically relevant surface geometry under the evaluated acquisition conditions.
In Fig.~\ref{fig:3d_recon_real_samples}, we present a few examples of 3D reconstructions from clinical data.

\vspace{-6pt}
\subsection{3D segmentation evaluation}

Three-dimensional segmentation was evaluated using the prospective dataset because it contains the multi-view information required by the aggregation method. Reference views were held out from aggregation and annotated by wound experts. The aggregated 3D labels were reprojected into these views and compared with the expert annotations using the Dice Similarity Coefficient (DSC). The corresponding direct 2D predictions were used as the paired baseline. Necrotic and epithelial tissue were excluded from inferential analysis because only 3 and 2 annotated images, respectively, were available.

\begin{figure}[!t]
    \centering
    \includegraphics[width=\linewidth]{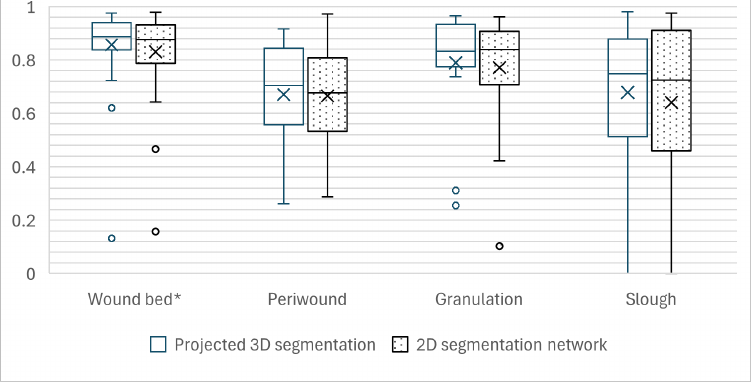}
    \caption{Distribution of the DSC for wound bed, periwound, granulation, and slough segmentation on the held-out prospective reference views. Necrotic and Epithelial tissue were excluded due to a lack of data. The asterisk denotes a statistically significant difference. `X' represents the mean.}
    \label{fig:dsc_stats}
\vspace{-6pt}
\end{figure}

The 2D-to-3D method performed better than or equal to the 2D method under all segmentation classes, as we observe in Fig.~\ref{fig:dsc_stats}.
%
A Wilcoxon signed-rank test identified a statistically significant difference only for wound bed segmentation ($p=0.00075$). 
Granulation did not reach statistical significance ($p=0.080$), potentially due to the smaller sample size ($N=23$). 
Furthermore, due to an insufficient sample size ($N=3$), periwound and slough classes could not provide meaningful statistical insights.
%
Overall, these results indicate that multi-view aggregation can improve wound-bed consistency and may mitigate isolated single-view errors. However, non-significant differences should not be interpreted as equivalent performance, and systematic errors in the underlying 2D predictions can propagate into the aggregated result.

\begin{figure}[!t]
    \centering     
    \includegraphics[width=\linewidth]{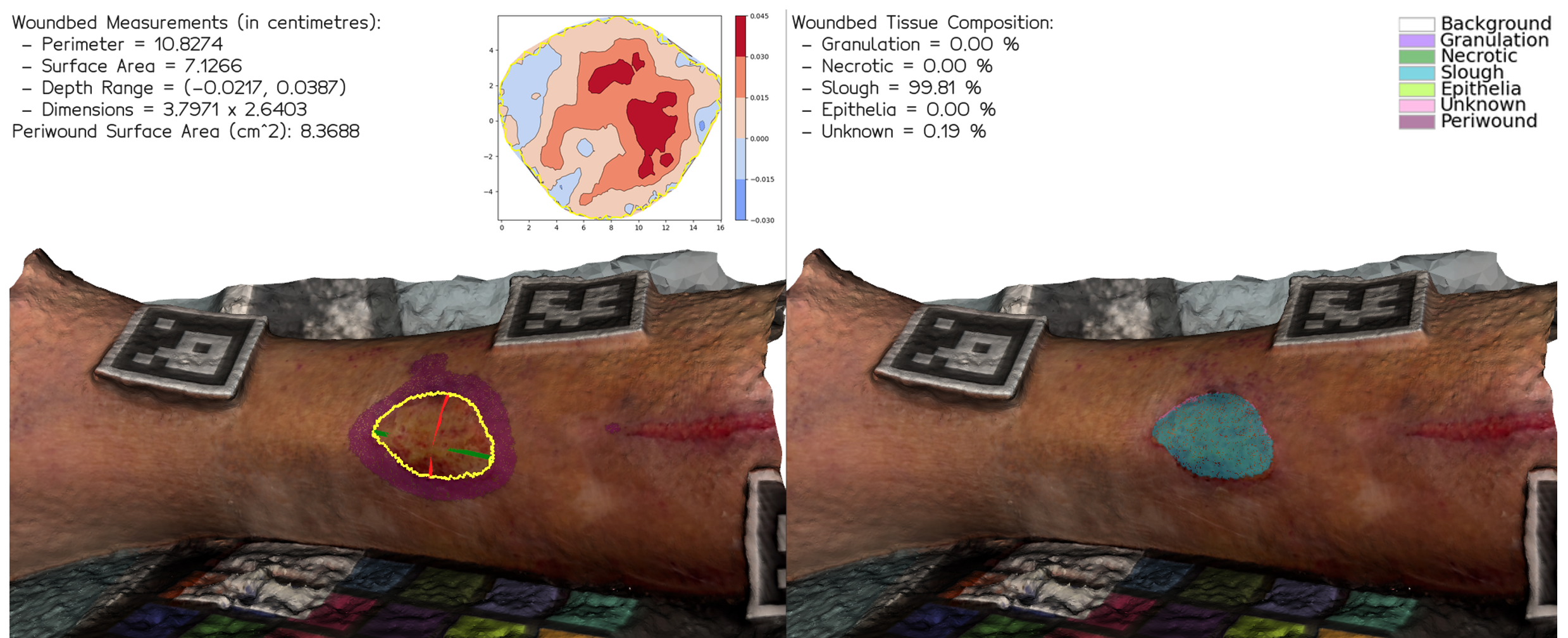} \\
    \includegraphics[width=\linewidth]{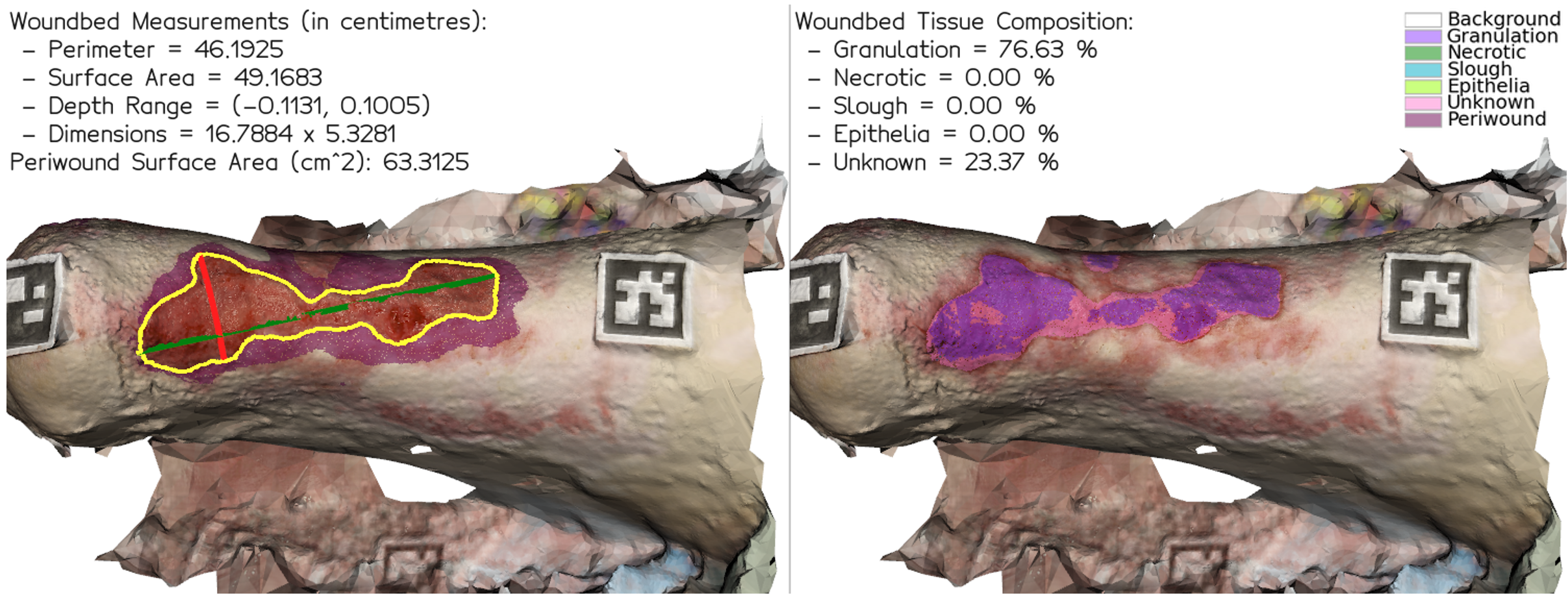}
    \caption{Wound documentation metrics generated by our framework from selected video recordings: \texttt{P3-01} and \texttt{P4-01}. The outputs include 3D wound geometry, wound bed measurements, wound bed tissue composition, and periwound area.}
    \label{fig:sample_wound_measurement}
\vspace{-6pt}
\end{figure}

\begin{figure}[!t]
    \centering
    \includegraphics[width=\linewidth]{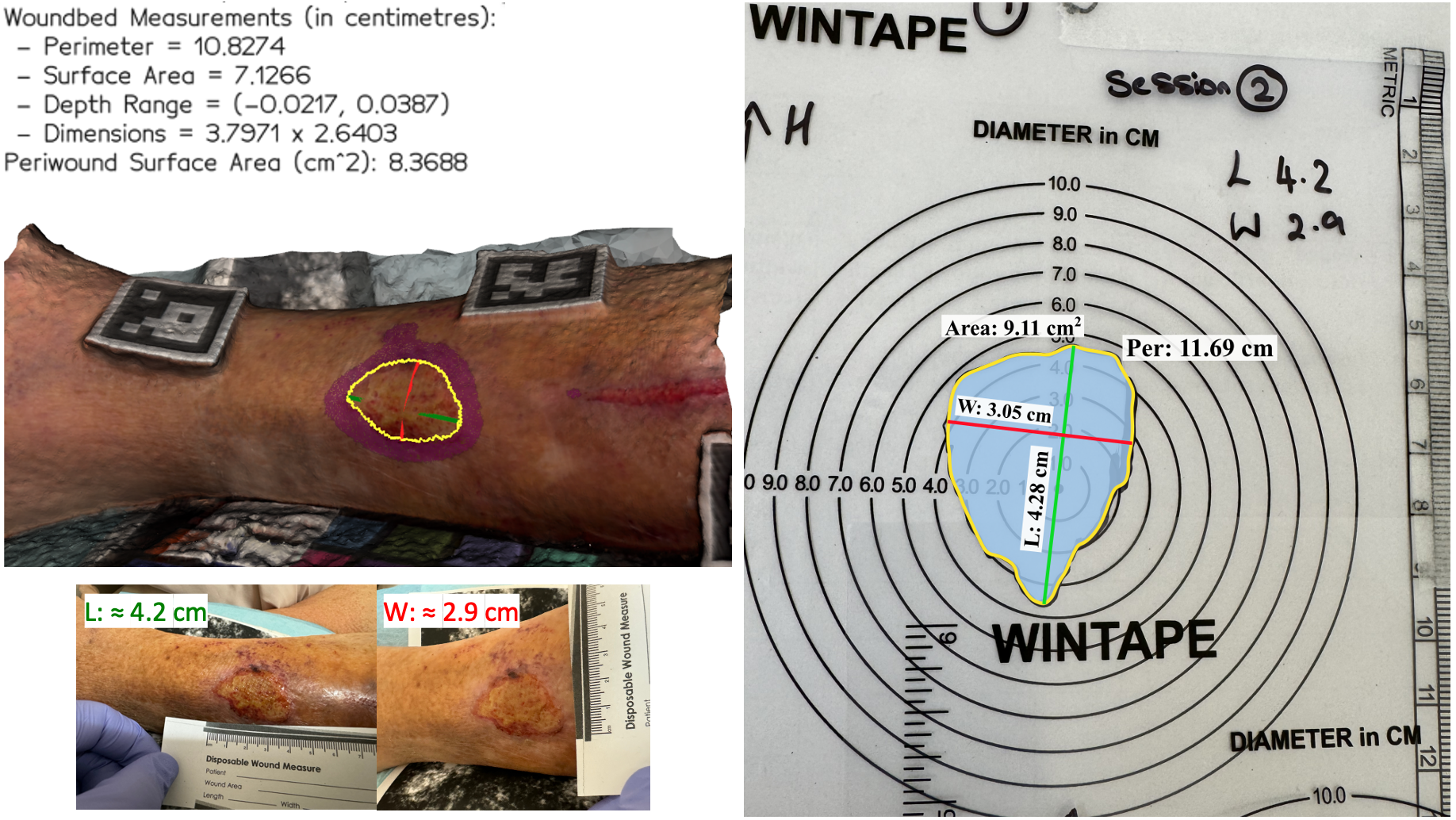}
    \caption{Comparison of framework-generated and manual documentation measurements for the wound \texttt{P3-01}.}
    \label{fig:wound_doc_manual_meas}
\vspace{-6pt}
\end{figure}

\begin{figure*}[!t]
    \centering
    \begin{overpic}[width=0.9\linewidth]{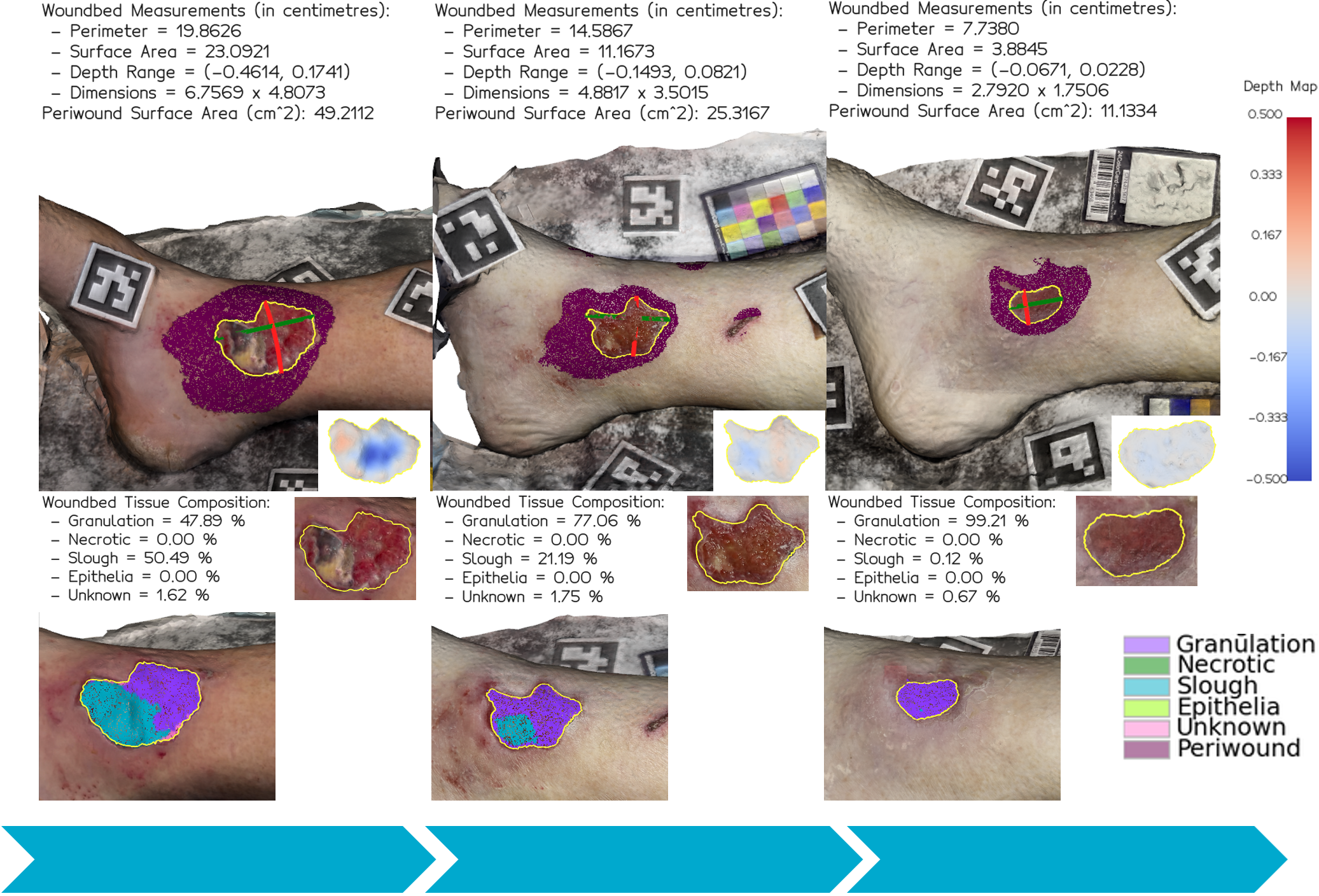} 
        \put (160,20) {\small \textcolor{white}{$T_0$}}
        \put (440,20) {\small \textcolor{white}{$T_0+2$ months}}
        \put (760,20) {\small \textcolor{white}{$T_0+3$ months}}
    \end{overpic}
    \caption{Longitudinal wound documentation results across consultations for patient \texttt{P5}. The figure shows wound bed measurements, wound bed tissue composition, and the appearance of the wounds on each consultation for a trajectory with decreasing wound area.}
    \label{fig:longitudinal_tracking_1001}
\vspace{-6pt}
\end{figure*}

\vspace{-6pt}
\subsection{Wound bed documentation: prospective data}

As described in Section~\ref{sec:woundbed_model}, our framework automatically computes clinically relevant wound bed metrics, including area, perimeter, dimensions, depth, and tissue composition. 
Fig.~\ref{fig:sample_wound_measurement} displays examples of the framework's output from real wound video recordings.
We compared these outputs with Wintape traces and ruler-based measurements used in routine documentation~\cite{haghpanah2006reliability}.
However, such comparisons are inherently limited due to subjectivity, lack of depth capture, and coarse manual methods. For instance, Wintape fails to account for cavity depth, and ruler-based approaches are highly dependent on user interpretation.
Nevertheless, these manual measurements were employed as coarse benchmarks to ensure consistency and to sanity-check our framework’s outputs. Additionally, we evaluated the longitudinal consistency of documentation outputs across sessions, providing a qualitative check on their alignment with visual wound progression.

\subsubsection{Comparison of wound documentation methods}

Fig.~\ref{fig:wound_doc_manual_meas} illustrates a comparison between manual clinician-reported measurements and framework-generated metrics for a representative case.
Manual methods, especially ruler-based estimation, can show variability across sessions, particularly in cases where body curvature affects the delineation of wound boundaries~\cite{haghpanah2006reliability}. While widely used in clinical practice, these techniques may have limitations in capturing complex wound geometries or depth.
Wintape traces, though more structured, tend to oversimplify wound geometry and do not reflect depth or cavity complexity.
Compared to these baselines, the framework-generated measurements were broadly consistent with the size and shape represented by the manual measurements in the illustrated cases. 
Because precise depth and volume measurements are clinically infeasible to obtain manually, we relied on the 3D scanner evaluations (Section~\ref{sec:eval_3d_proc}) to assess geometric accuracy.
However, these comparisons evaluate overall surface geometry rather than directly validating the derived maximum-depth or volume measurements.
While this means the clinical accuracy of each 3D metric is not yet fully established, these results demonstrate the framework’s ability to generate rich, non-contact geometric descriptors that remain unattainable through standard manual techniques.

\subsubsection{Longitudinal tracking of wound documentation}

We also explored the use of our framework for tracking wound progression across multiple visits in a subset of patients.

Fig.~\ref{fig:longitudinal_tracking_1001} presents results from \texttt{P5}. 
Over approximately three months, the wound showed clear signs of healing: the wound bed area decreased from $23.09~\text{cm}^2$ to $3.88~\text{cm}^2$, and slough tissue transitioned almost entirely to granulation tissue, following trends consistent with expected healing stages.

\begin{figure*}[!t]
    \centering
    \begin{overpic}[width=0.9\linewidth]{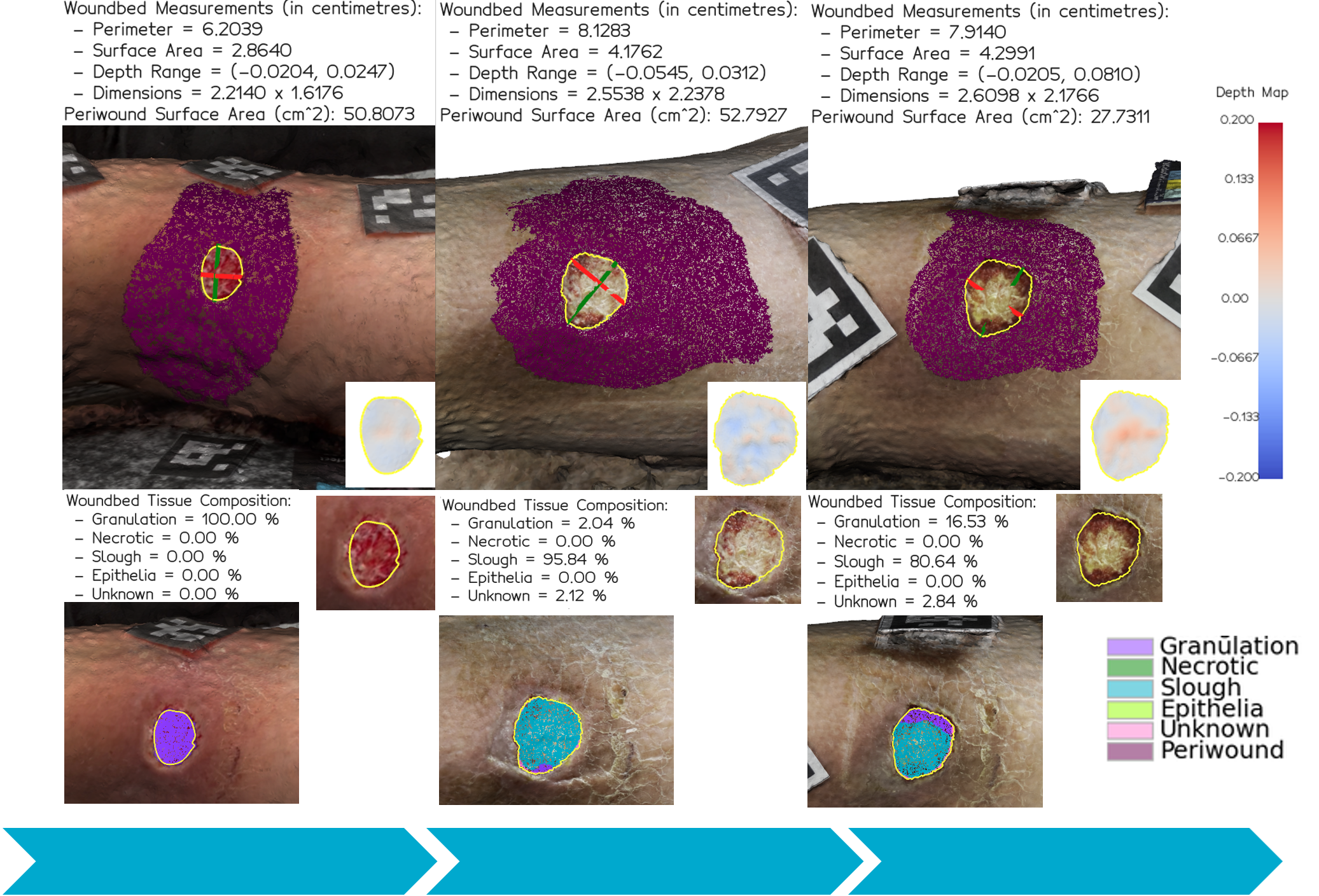} 
        \put (160,20) {\small \textcolor{white}{$T_0$}}
        \put (440,20) {\small \textcolor{white}{$T_0+2$ months}}
        \put (760,20) {\small \textcolor{white}{$T_0+3$ months}}
    \end{overpic}
    \caption{Longitudinal wound documentation results across consultations for patient \texttt{P6}. The figure shows wound bed measurements, wound bed tissue composition, and the appearance of the wounds at each consultation for a trajectory with increasing wound area.}
    \label{fig:longitudinal_tracking_1002}
\vspace{-6pt}
\end{figure*}

In contrast, Fig.~\ref{fig:longitudinal_tracking_1002} shows data from \texttt{P6}, where the wound area increased from $2.9~\text{cm}^2$ to $4.3~\text{cm}^2$ over a similar period.
Tissue composition showed irregular changes, including a rise and subsequent drop in slough presence. 
Identifying these patterns provides an early indicator of impaired healing, enabling clinicians to reassess treatment strategies promptly.

Our prospective dataset does not include large-scale expert annotations of exact measurements, which limits the extent of quantitative benchmarking. However, the combination of clinical comparisons, temporal tracking, and consistency with expected wound progression patterns provides early and compelling evidence of Wound3DAssist’s reliability. Future work will prioritize expanding annotated datasets and formalizing expert assessment to strengthen clinical confidence.

\vspace{-6pt}
\subsection{Framework performance and hardware requirements}

\begin{figure*}[!h]
     \centering
     \includegraphics[width=\linewidth]{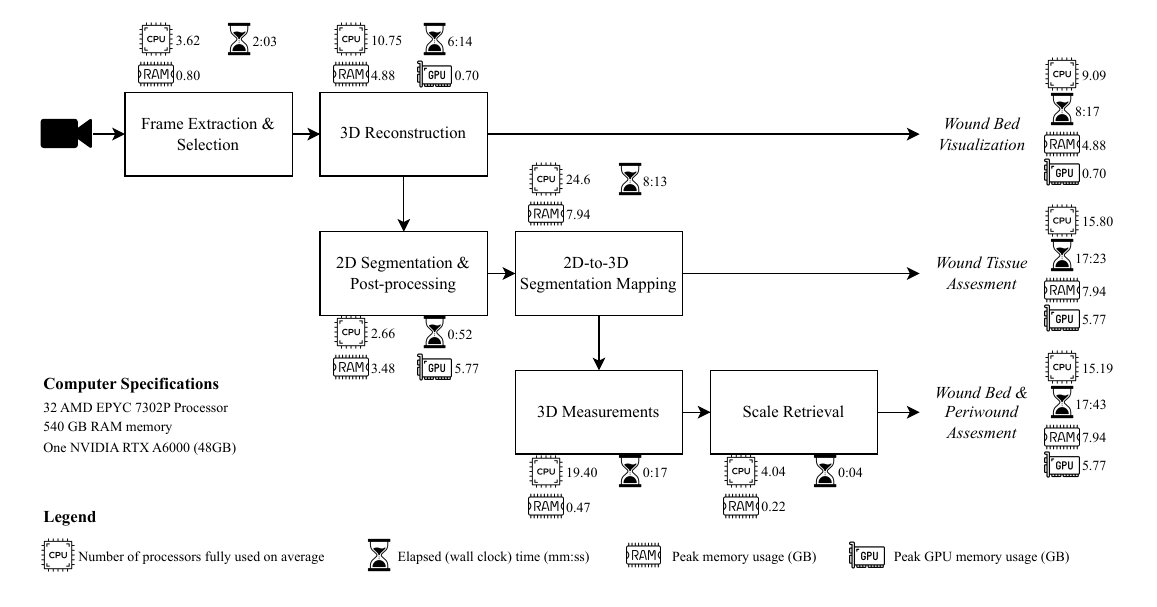}
     \caption{Runtime and hardware usage across framework components, averaged over the prospective validation dataset.}
     \label{fig:hardware_full}
 \vspace{-6pt}
 \end{figure*}

We evaluated the computational performance and hardware requirements of the Wound3DAssist framework.
All experiments were conducted on a high-performance computing setup featuring a 32-core AMD EPYC 7302P processor, 540 GB of RAM, and an NVIDIA RTX A6000 GPU (48 GB).
The full documentation pipeline is completed in approximately 18 minutes.
Initial 3D visualization is available within 8 minutes, while tissue assessment takes an additional 10 minutes for 2D segmentation and 2D-to-3D mapping. Wound measurements are computed in a matter of seconds. 
Resource usage remains moderate, with peak GPU and RAM consumption below 8GB, making the framework viable for consumer-grade hardware. 
The most time-consuming components are 3D reconstruction and 2D-3D segmentation, indicating opportunities for future optimization.
In Fig.~\ref{fig:hardware_full} we present a detailed breakdown of processing time and resource usage for the key components.

\begin{figure*}[!t]
    \centering
    \includegraphics[width=0.9\linewidth]{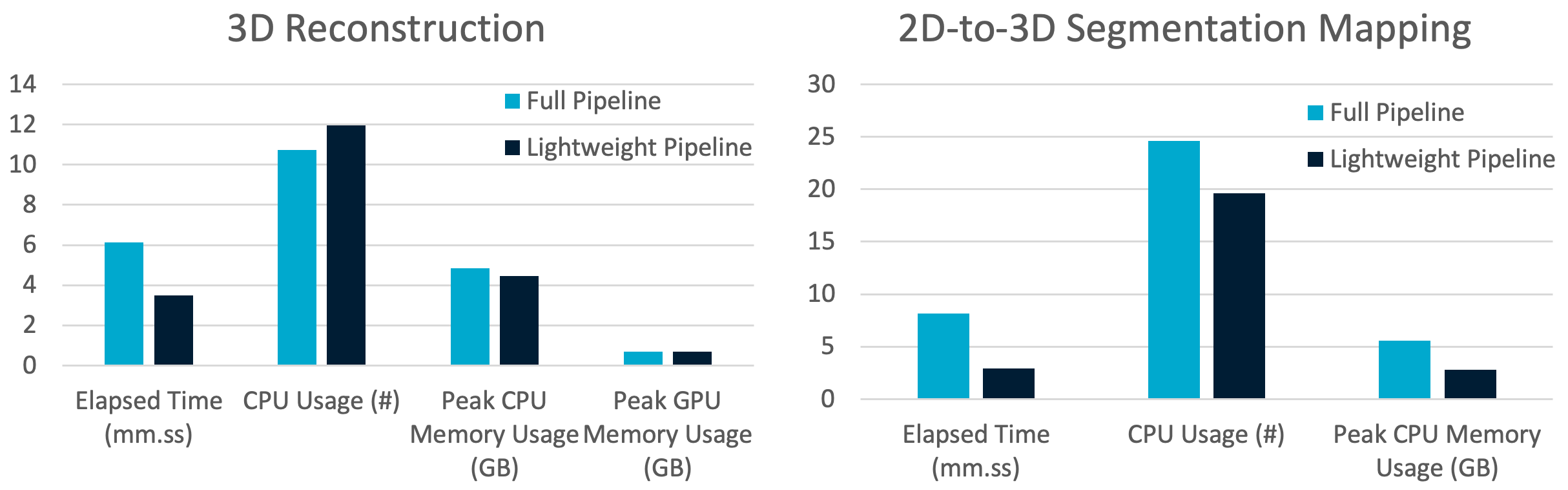}
    \caption{Comparison of the full and lightweight pipelines in terms of elapsed time, CPU usage, RAM, and GPU memory for the two most impacted stages---3D reconstruction and 2D-to-3D segmentation mapping.}
    \label{fig:hardware_fast}
\vspace{-6pt}
\end{figure*}

To improve speed, we also developed a lightweight version of the Wound3DAssist framework. 
This variant employs a simplified 3D reconstruction pipeline that bypasses dense reconstruction, generating meshes from the sparse point cloud. This substantially reduces the number of mesh faces, from hundreds of thousands to tens of thousands, accelerating both reconstruction and segmentation. 
Fig.~\ref{fig:hardware_fast} illustrates the gains achieved with this streamlined pipeline.
The lightweight configuration reduces reconstruction and mapping time by generating a lower-complexity mesh from the sparse point cloud. However, it may omit fine wound-bed geometry. These results demonstrate a configurable speed–fidelity trade-off rather than readiness for a particular clinical workflow. Further evaluation is required to determine acceptable processing times and geometric fidelity in specific care settings.


\section{Discussion}
We validated the technical feasibility of our wound documentation framework, \textit{Wound3DAssist}, across three datasets of increasing complexity, focusing on surface reconstruction, segmentation, wound documentation, longitudinal assessment, and computational efficiency.

%
The digital dataset provided known geometry, while the silicone dataset, detailed in the SALVE benchmark~\cite{chierchia2024salve}, enabled controlled evaluation of reconstruction precision, device variability, and acquisition artefacts, an assessment often impractical in real clinical scenarios. 
SALVE reported sub-millimeter precision for selected reconstruction methods under controlled silicone-wound conditions.
In the clinical dataset, Wound3DAssist achieved an average point-wise surface error of $1~\text{mm}$ over four real-world cases. 
This result reflects overall surface-reconstruction fidelity and should not be interpreted as direct validation of maximum wound depth or volume.
Although Filko et al.~\cite{filko2021automatic} reported a mean accuracy of $0.14~\text{mm}$ using a robotic arm with a high-end 3D scanner, their system requires specialised, non-portable hardware.

On the prospective clinical dataset, the 2D-to-3D method consistently outperformed direct 2D methods, with improved Dice Similarity Coefficient for wound bed, periwound, and two key tissue types: granulation and slough.
These findings suggest that multi-view aggregation can improve wound-bed consistency and mitigate isolated single-view errors, although systematic errors in the underlying 2D predictions may propagate into the aggregated result. This is consistent with prior findings that emphasize the advantage of multi-view analysis in medical image segmentation~\cite{niri2021multi}.

The longitudinal examples further demonstrated the framework's ability to quantify changes across assessments.
For patient \texttt{P5}, the framework captured reductions in surface area and depth, metrics linked to healing trajectories~\cite{flanagan2003wound}. 
On the contrary, patient \texttt{P6} exhibited non-healing progression, which our system successfully quantified. 
While these metrics can highlight changes that warrant clinical review, they do not constitute an automated diagnostic classification; rather, these results suggest Wound3DAssist can provide clinicians with an objective, reproducible tool for monitoring wound progression.

Unlike traditional clinical practices that are labor-intensive and constrained by a sterile environment, our framework enables non-invasive documentation using widespread smartphone cameras. 
This facilitates more frequent reassessment of treatment, beyond the typical recommendation of two to four weeks~\cite{flanagan2003wound}.
While commercial solutions exist~\footnote{https://www.aranzmedical.com/}~\footnote{https://ekare.ai/}, they rely on proprietary hardware and manual segmentation, limiting routine clinical integration.
The complete pipeline required approximately 18 minutes on a high-performance workstation. A lightweight variant reduced processing time at the cost of surface detail and reconstruction completeness. The modular design also permits future integration of alternative reconstruction methods; however, neural approaches that improved surface fidelity in selected benchmarking conditions required greater computational resources and processing time.

Several acquisition-related failure modes remain. Severe motion blur can prevent reliable feature matching; reflective or moist surfaces may introduce erroneous geometry; and occlusions reduce visible surface coverage. Failure to capture an ArUco marker prevents metric scale recovery, while deep, undermined, or tunnelling structures cannot be reconstructed when they are not visible in the RGB views. Frame selection reduces, but does not eliminate, these limitations.

The clinical reconstruction evaluation was limited to four manually aligned cases, and direct reference measurements of maximum depth, volume, and dense 3D segmentation were unavailable. The small prospective cohort prevented reliable subgroup analyses by wound aetiology, anatomical location, wound size, skin tone, lighting, or occlusion, while single-expert annotations prevented assessment of inter-rater reliability. The periwound component is limited to spatial delineation and surface-area estimation and does not assess diagnostic characteristics such as erythema, oedema, infection, or temperature. Moreover, the prospective cohort is an unseen test set rather than an external or multi-centre validation cohort. Wound3DAssist should therefore be considered a research prototype requiring larger independent validation and formal usability and workflow studies.
Future work will focus on expanding the clinical dataset, particularly for segmentation annotations, potentially by leveraging pseudo-labeling and generative data augmentation~\cite{koohi2023generative}.
We will also explore recent neural rendering variants capable of learning from blur artifacts~\cite{Ma_2022_CVPR} to improve robustness without sacrificing runtime efficiency. 
These enhancements aim to build a foundation for practical, scalable wound documentation in clinical environments.

\section{Conclusion}

We introduce Wound3DAssist, a practical and modular framework for 3D wound assessment using monocular videos captured with consumer-grade devices.
Through extensive evaluation across synthetic, phantom, and clinical wound data, we demonstrated its ability to generate accurate 3D reconstructions, extract key wound metrics, and enable consistent longitudinal tracking. 
The framework supports interoperable, end-to-end analysis, including tissue classification and periwound assessment, while maintaining efficient processing, with full assessment completed in 20 minutes.
Its modular design allows for seamless integration of future advances, offering a scalable and cost-effective solution for modern wound care.

\section*{Acknowledgements}
The authors would like to acknowledge Hayley Ryan and WoundRescue Pty Ltd who supported development of the retrospective clinical wound dataset, collection of wound images and image annotations, Prof. Michelle Barakat-Johnson who supported the development of the synthetic dataset, and Ms Joanne Marjoram and Ms Cassandra Kelly, clinical nurse researchers at the University of Sydney School of Rural Health, for facilitating and conducting the collection of chronic wound images for the prospective data. We also thank A/Prof. Georgina Luscombe for leading the ethical approvals, data assessment and validation, and the review of user requirements; Ms Kate Smith for research ethics and governance support; and Dr Annie Banbury and Ms Melanie Pefani, Coviu Global Pty Ltd., for assistance with image annotation and the development and review of user requirements.

\section*{Compliance with Ethical Standards}
This study was performed in line with the principles of the Declaration of Helsinki. The experimental procedures involving human subjects described in this paper
were approved by the CSIRO Health and Medical Human Research Ethics Committee and the University of Sydney, and the Greater Western Human Research Ethics Committees 
[Ethics protocol number: 2022/HE000523; 2022/HE000820; 2023/ETH01164; 2023/HE000916].

\bibliographystyle{IEEEtran}
{\footnotesize
\bibliography{refs}}

\end{document}

%% file: tables/system_comparison.tex
\begin{table*}[!ht]
\centering
\caption{Comparison of our proposed system for 3D wound analysis with other related works using photogrammetry and consumer-grade devices.}
    \begin{tabular}{lccccccc}\toprule
    
    \multicolumn{1}{c}{} &\multicolumn{1}{c}{~\cite{wannous2010enhanced}} &\multicolumn{1}{c}{~\cite{mirzaalian2019measuring}} &\multicolumn{1}{c}{~\cite{liu2019wound}} &\multicolumn{1}{c}{~\cite{sanchez2022sfm}} &\multicolumn{1}{c}{~\cite{souto2023three}} &\multicolumn{1}{c}{~\cite{niri2021multi}} &\multicolumn{1}{c}{Ours} \\
    \midrule
    
    Recent photogrammetry method    &   &   &   &   &\y &\y &\y \\
    Recent segmentation method      &   &   &   &   &\y &\y &\y \\
    Automatic wound segmentation    &\y &   &   &   &   &\y &\y \\
    Wound multi-class tissue segmentation       &\y &\y &   &   &\y &   &\y \\
    Wound depth assessment   &\y &   &   &\y &   &   &\y \\
    \bottomrule
    \end{tabular}
\vspace{-8pt}
\label{tab:system_comp}
\end{table*}